\documentclass[twoside]{article}
\usepackage[accepted]{aistats2018}
\usepackage{natbib}
\usepackage{graphicx} 
\usepackage{subfigure}
\usepackage[utf8]{inputenc} 
\usepackage[T1]{fontenc}    
\usepackage{hyperref}       
\usepackage{url}            
\usepackage{booktabs}       
\usepackage{amsfonts}       
\usepackage{nicefrac}       
\usepackage{microtype}      
\usepackage{amsmath}
\usepackage{multirow}

\newcommand{\comment}[1]  {}
\usepackage{color}

\graphicspath{ {figures/} }

\begin{document}

%

%

\twocolumn[

\aistatstitle{Distance-based Confidence Score for Neural Network Classifiers}

\aistatsauthor{Amit Mandelbaum \And Daphna Weinshall}

\aistatsaddress{School of Computer Science and Engineering \\ Hebrew University of Jerusalem, Israel \And School of Computer Science and Engineering \\ Hebrew University of Jerusalem, Israel} 
]

\begin{abstract}
The reliable measurement of confidence in classifiers' predictions is very important for many applications, and is therefore an important part of classifier design. Yet, although deep learning has received tremendous attention in recent years, not much progress has been made in quantifying the prediction confidence of  neural network classifiers. Bayesian models offer a mathematically grounded framework to reason about model uncertainty, but usually come with prohibitive computational costs. In this paper we propose a simple, scalable method to achieve a reliable confidence score, based on the data embedding derived from the penultimate layer of the network. We investigate two ways to achieve desirable embeddings, by using either a distance-based loss or Adversarial Training. We then test the benefits of our method when used for classification error prediction, weighting an ensemble of classifiers, and novelty detection. In all tasks we show significant improvement over traditional, commonly used confidence scores. 
\end{abstract}

\section{Introduction}
\label{sec:intro}

Classification confidence scores are designed to measure the accuracy of the model when predicting class assignment (rather than the uncertainty inherent in the data). Most generative classification models are probabilistic in nature, and therefore provide such confidence scores directly. Most discriminative models, on the other hand, do not have direct access to the probability of each prediction. Instead, related non-probabilistic scores are used as proxies, as for example the margin in SVM classifiers. 

When trying to evaluate the confidence of neural network (NN) classifiers, a number of scores are commonly used. One is the strength of the most activated output unit followed by softmax normalization, or the closely related ratio between the activities of the strongest and second strongest units. Another is the (negative) entropy of the output units, which is minimal when all units are equally probable. Often, however, these scores do not provide a reliable measure of confidence. 

Why is it important to reliably measure prediction confidence? In various contexts such as medical diagnosis and decision support systems, it is important to know the prediction confidence in order to decide how to act upon it. For example, if the confidence in a certain prediction is too low, the involvement of a human expert in the decision process may be called for. Another important aspect of real world applications is the ability to recognize samples that do not belong to any of the known classes, which can also be improved with a reliable confidence score.  But even irrespective of the application context, reliable prediction confidence can be used to boost the classifier performance via such methods as self-training or ensemble classification. In this context a better confidence score can improve the final performance of the classifier. The derivation of a good confidence score should therefore be part of the classifier's design, as important as any other component of classifiers' design.

In order to derive a reliable confidence score for NN classifiers, we focus our attention on an empirical observation concerning neural networks trained for classification, which have been shown to demonstrate in parallel useful embedding properties. Specifically, a common practice these days is to treat one of the upstream layers of a pre-trained network as a representation (or embedding) layer. This layer activation is then used for representing similar objects and train simpler classifiers (such as SVM, or shallower NNs) to perform different tasks, related but not identical to the original task the network had been trained on. 

In computer vision such embeddings are commonly obtained by training a deep network on the recognition of a very large database (typically ImageNet \citep{imagenet_cvpr09}). These embeddings have been shown to provide better semantic representations of images (as compared to more traditional image features) in a number of related tasks, including the classification of small datasets \citep{sharif2014cnn}, image annotation \citep{donahue2015long} and structured predictions \citep{hu2016learning}. Given this semantic representation, one can compute a natural multi-class probability distribution as described in Section~\ref{sec:new-conf-score}, by estimating local density in the embedding space. This estimated density can be used to assign a confidence score to each test point, using its likelihood to belong to the assigned class. 

We note, however, that the commonly used embedding discussed above is associated with a network trained for classification only, which may impede its suitability to measure confidence reliably. In fact, when training neural networks, metric learning is often used to achieve desirable embeddings (e.g., \cite{weston2012deep,schroff2015facenet,hoffer2015deep,tadmor2016learning}). Since our goal is to improve the probabilistic interpretation of the embedding, which is essentially based on local point density estimation (or the distance between points), we may wish to modify the loss function and add a term which penalizes for the violation of pairwise constraints as in \cite{hadsell2006dimensionality}. Our experiments show that the modified network indeed produces a better confidence score, with comparable classification performance. Surprisingly, while not directly designed for this purpose, we show that networks which are trained with adversarial examples following the Adversarial Training paradigm \citep{szegedy2013intriguing,goodfellow2014explaining}, also provide a suitable embedding for the new confidence score. 

Our first contribution, therefore, is a new prediction confidence score which is based on local density estimation in the embedding space of the neural network. This score can be computed for every network, but in order for this score to achieve superior performance, it is necessary to slightly change the training procedure. In our second contribution we show that suitable embedding can be achieved by either augmenting the loss function of the trained network with a term which penalizes for distance-based similarity loss (as in Eq.~(\ref{eq:dist-loss}) below), or by using Adversarial Training. The importance of the latter contribution is two fold: Firstly, we are the first to show that the density of image embeddings is improved with indirect Adversarial Training perturbations, in addition to the improved word embedding quality shown in \cite{miyato2016adversarial} by direct Adversarial Training perturbations. Secondly, we show in Section~\ref{sec:exp} that Adversarial Training improves the results while imposing a much lighter burden of hyper-parameters to tune as compared to the distance-based loss.

The new confidence score is evaluated in comparison to other scores, using the following tasks: (i) Performance in the binary classification task of identifying each class prediction as correct or incorrect (see Section~\ref{sec:new-conf-score}).
(ii) Training an ensemble of NN classifiers, where each classifier's prediction is weighted by the new confidence score (see Section~\ref{sec:new-ensemble}). (iii) Novelty detection, where confidence is used to predict whether a test point belongs to one of the known classes from the train set (see Section~\ref{sec:new-novelty}). 

The empirical evaluation of our method is described in Section~\ref{sec:exp}, using a few datasets and different network architectures which have been used in previous work when using these specific datasets. Our method achieves significant improvement in all 3 tasks. When compared with a more recent method which had been shown to improve traditional measures of classification confidence - MC dropout \citep{gal2015dropout}, our distance-based score achieves better results while also maintaining lower computational costs. 

\subsection*{Prior Work} 
\label{sec:prior}

\comment{\paragraph*{Confidence score for neural network classifiers:} }The
Bayesian approach seeks to compute a posterior distribution over the parameters of the neural network which is used to estimate prediction uncertainty, as in \cite{mackay1992bayesian} and \cite{neal2012bayesian}. However, Bayesian neural networks are not always practical to implement, and the computational cost  involved it typically high. 
In accordance, in a method which is referred to below as \emph{MC-Dropout}, \cite{gal2015dropout} proposed to use dropout during test time as a Bayesian approximation of the neural network, providing a cheap proxy to Bayesian Neural Networks. \cite{lakshminarayanan2016simple} proposed to use Adversarial Training 
to improve the uncertainty measure of the entropy score of the neural network.

Still, the most basic and one of the most common confidence scores for neural networks can be derived from the strength of the most activated output unit, \comment{(also called \textit{softmax input}). In doing so we assume that our model is reasonably good in approximating the real class posterior probabilities:
\begin{equation}
D_0(x) = max_k\hat{p}(C=k|x)
\end{equation}
Note that the network's output, $\hat{p}(k|x)$, are not however true probabilities, since there is no guarantee that their sum be one. A probability distribution may be trivially obtained by normalizing the outputs. A more statistically meaningful score would therefore be the} or rather its normalized version (also called \textit{softmax output} or \emph{max margin})\comment{:
\begin{equation}
D_1(x) = max_k\frac{\hat{p}(k|x)}{\sum_{j=1}^{m}\hat{p}(j|x)}
\end{equation}
However, these scores are an approximation of the true probability of good classification when the class chosen is most probably correct, but not otherwise. They do not tell us the confidence on the choice of class. 

}. A confidence score that handles better a situation where there is no one class which is most probable, is the (negative) entropy of the normalized network's output.\comment{ \cite{wan1990neural}:
\begin{equation}
D_2(x) =-H(p(\hat{k}|x))=\sum_{j=1}^{m}p(\hat{j}|x)\log p(\hat{j}|x)
\end{equation}
Where $p(\hat{j}|x)$ equals to $D_1(x)$.} \cite{zaragoza1998confidence} compared these scores, as well as some more complex ones (e.g. \cite{tibshirani1996comparison}), demonstrating somewhat surprisingly the empirical superiority of the two most basic methods described in the previous paragraph. 



Ensembles of models have been used to improve the overall performance of the final classifier (see reviews in  \cite{dietterich2000ensemble} and \cite{li2017research}). There are many ways to train an ensemble, such as boosting or bagging.\comment{, in the context of this work we address the kind of ensemble which is made when using different training parameters with a single training method. 

The simplest and most common method for using the ensemble mentioned above is by averaging the prediction of all classifier into a single prediction. Another option commonly used is some sort of voting mechanism simple or more complexed } There are also many ways to integrate the predictions of the classifiers in the ensemble, including the average prediction or voting discussed by \cite{bauer1999empirical}. Some ensemble methods use the confidence score to either weight the predictions of the different classifiers (average weighting) or for confidence voting\comment{ where the most confident networks gets more (or even all) votes when classifying}. 

Novelty detection, where the task is to determine whether a test point belongs to a known class label or not, is another problem which becomes more relevant with the ever increasing availability of very large datasets, see reviews in \cite{markou2003novelty}, \cite{pimentel2014review} and the recent work in \cite{vinokurov2016novelty}. This task is also highly relevant in real world applications, where the classifier is usually exposed to many samples which do not belong to a known class.
Note that novelty \emph{detection} is quite different from the learning of classes with no examples, as in zero shot learning \citep{palatucci2009zero}.

\section{New Confidence Score} 
\label{sec:conf-score}

We propose next a new confidence score. We then discuss how it can be used to boost classification performance with ensemble methods, or when dealing with novelty detection.

\subsection{New Confidence Score for Neural Network Classifiers}
\label{sec:new-conf-score}


Our confidence score is based on the estimation of local density as induced by the network, when points are represented using the effective embedding created by the trained network in one of its upstream layers. Local density at a point is estimated based on the Euclidean distance in the embedded space between the point and its $k$ nearest neighbors in the training set. 

Specifically, let $f(x)$ denote the embedding of $x$ as defined by the trained neural network classifier. Let $A(x)=\{x_{train}^j\}_{j=1}^k$ denote the set of $k$-nearest neighbors of $x$ in the training set, based on the Euclidean distance in the embedded space, and let $\{y^j\}_{j=1}^k$ denote the corresponding class labels of the points in $A(x)$. A probability space is constructed (as is customary) by assuming that the likelihood that two points belong to the same class is proportional to the exponential of the negative Euclidean distance between them. In accordance, the local probability that a point $x$ belongs to class $c$ is proportional to the probability that it belongs to the same class as the subset of points in $A(x)$ that belong to class $c$. 

Based on this local probability, the confidence score $D(x)$ for the assignment of point $x$ to class $\hat{y}$ is defined as follows:   
\begin{equation}
D(x) = \frac{\sum_{j=1, y^{j}=\hat{y}}^{k}e^{-||f(x)-f(x_{train}^{j})||_2}}{\sum_{j=1}^{k}e^{-||f(x)-f(x_{train}^{j})||_2}}
\label{eq:distscore}
\end{equation}
$D(x)$ is a score between 0 to 1, which is monotonically related to the local density of similarly labeled train points in the neighborhood of $x$. Henceforth (\ref{eq:distscore}) is  referred to as \emph{Distance score}.\footnote{Related measures of density, such as a count of the "correct" neighbors or the inverse of the distance, behave similarly and perform comparably.} We note here that while intuitively it might be beneficial to add a scaling factor to the distance in  (\ref{eq:distscore}), such as the mean distance, we found it to have a deteriorating effect, in line with related work such as \cite{salakhutdinov2007learning}.

\paragraph{Two ways to achieve effective embedding:} As mentioned is Section~\ref{sec:intro}, in order to achieve an effective embedding it helps to modify the training procedure of the neural network classifier. The simplest modification  augments the network's loss function during training with an additional term. The resulting loss function is a linear combination of two terms, one for classification denoted $\mathcal{L}_{class}(X,Y)$, and another pairwise loss for the embedding denoted $\mathcal{L}_{dist}(X,Y)$. This is defined as follows: 
\begin{align}
\label{eq:dist-loss}
&\mathcal{L}(X,Y)=\mathcal{L}_{class}(X,Y)+\alpha\mathcal{L}_{dist}(X,Y) \\
&\mathcal{L}_{dist}(X,Y) = \frac{1}{P}\sum_{p=1}^{P}{L}_{dist}(x^{p_1},x^{p_2}) \nonumber
\end{align}
where ${L}_{dist}(x^{i},x^{j})$ is defined as
\begin{equation*}
\begin{cases} ||f(x^{i})-f(x^{j})||_2 & \text{if } y^{i}=y^{j}\\  max\{0,(m-||f(x^{i})-f(x^{j})||_2)\} & \text{if }  y^{i}\neq y^{j}
\end{cases}
\end{equation*}



A desirable embedding can also be achieved by Adversarial Training, using the \textit{fast gradient method} suggested in \cite{goodfellow2014explaining}. In this method, given an input $x$ with target $y$, and a neural network with parameters $\theta$, adversarial examples are generated using: 
\begin{equation}
\label{eq:adverserial}
x' = x + \epsilon~ sign(\bigtriangledown _x L_{class}(\theta,x,y))
\end{equation}
In each step an adversarial example is generated for each point $x$ in the batch and the current parameters of the network, and classification loss is minimized for both the regular and adversarial examples. Although originally designed to improve robustness, this method seems to improve the network's embedding for the purpose of density estimation, possibly because along the way it increases the distance between pairs of adjacent points with different labels. 

\paragraph{Implementation details:} In (\ref{eq:dist-loss}) $\mathcal{L}_{dist}$ is defined by all pairs of points, denoted $(x^{p_1},x^{p_2})$. For each training minibatch, this set is sampled with no replacement from the training points in the minibatch, with half as many pairs as the size of the minibatch. In our experiments, $\mathcal{L}_{class}(X,Y)$ is the regular cross entropy loss. We note here that we also tried distance-based loss functions which do not limit the distance between points of the same class to be exactly 0 (such as those in \cite{hoffer2015deep} and \cite{tadmor2016learning}). However, those functions produced worse results, especially when the dataset had many classes. Finally we note that we have tried using the distance-based loss and adversarial training together while training the network, but this has also produced worse results.

\subsection{Alternative confidence scores} 
\label{sec:otherconf}
Given a trained network, two measure are usually used to evaluate classification confidence: 
\begin{description}
\item[Max margin:]
the maximal activation, after normalization, in the output layer of the network.
\item[Entropy:]
the (negative) entropy of the activations in the output layer of the network.
\end{description}
As noted above, the empirical study in \cite{zaragoza1998confidence} showed that these two measures are typically as good as any other existing method for the evaluation of classification confidence.

Two recent methods have been shown to improve the reliability of the confidence score based on Entropy: \textit{MC-Dropout} \citep{gal2015dropout} and \textit{Adversarial Training}  \citep{lakshminarayanan2016simple,goodfellow2014explaining}. In terms of computational cost, adversarial training can increase (and sometimes double) the training time, due to the computation of additional gradients and the addition of the adversarial examples to the training set. MC-Dropout, on the other hand, does not change the training time but increases the test time by orders of magnitude (typically 100-fold). Both methods are complementary to our approach, in that they focus on modifications to the actual computation of the network during either train or test time. After all is done, they both evaluate confidence using the Entropy score. As we show in our experiments, adversarial training combined with our proposed confidence score improves the final results significantly.   

\subsection{Our method: computational analysis}

Unlike the two methods described above, \textit{MC-Dropout} and \textit{Adversarial Training}, our distance-based confidence score takes an existing network and computes a new confidence score from the network's embedding and output activation. It can use any network, with or without adversarial training or MC dropout. If the loss function of the network is suitably augmented (see discussion above), empirical results in Section~\ref{sec:exp} show that our score always improves results over the Entropy score of the given network. 

\paragraph{Train and test computational complexity:} 
Considering the distance-based loss, \cite{tadmor2016learning} showed that computing distances during the training of neural networks have negligible effect on training time. Alternatively, when using adversarial training, additional computational cost is incurred as mentioned above, while on the other hand fewer hyper parameters are left for tuning. During test time, our method requires carrying over the embeddings of the training data and also the computation of the $k$ nearest neighbors for each sample. 

Nearest neighbor classification has been studied extensively in the past 50 years, and consequently there are many methods to perform either precise or approximate $k$-nn with reduced time and space complexity (see \cite{gunadi2011comparing} for a recent empirical comparison of the main methods). In our experiments, while using either Condensed Nearest Neighbours \citep{hart1968condensed} or Density Preserving Sampling \citep{budka2013density}, we were able to reduce the memory requirements of the train set to $5\%$ of its original size without affecting performance. At this point the additional storage required for the nearest neighbor step was much smaller than the size of the networks used for classification, and the increase in space complexity became insignificant.

With regards to time complexity, recent studies have shown how modern GPU's can be used to speed up nearest neighbor computation by orders of magnitude \citep{garcia2008fast,arefin2012gpu}. \cite{hyvonen2015fast} also showed that k-nn approximation with 99\% recall can be accomplished 10-100 times faster as compared to precise k-nn. 

Combining such reductions in both space and time, we note that even for a very large dataset, including for example 1M images embedded in a 1K dimensional space, the computation complexity of the $k$ nearest neighbors for each test sample requires at most 5M floating-point operations. This is comparable and even much faster than a single forward run of this test sample through a modern, relatively small, ResNets \citep{he2016deep} with 2-30M parameters. Thus, our method scales well even for very large datasets.

\subsection{Ensembles of Classifiers}
\label{sec:new-ensemble}

There are many ways to define ensembles of classifiers, and different ways to put them together. Here we focus on ensembles which are obtained when using different training parameters with a single training method. This specifically means that we train several neural networks using random initialization of the network parameters, along with random shuffling of the train points.

Henceforth \emph{Regular Networks} will refer to networks that were trained only for classification with the regular cross-entropy loss, \emph{Distance Networks} will refer to networks that were trained with the loss function defined in (\ref{eq:dist-loss}), and \emph{AT Networks} will refer to networks that were trained with adversarial examples as defined in (\ref{eq:adverserial}).

Ensemble methods differ in how they weigh the predictions of different classifiers in the ensemble. A number of options are in common use (see \cite{li2017research} for a recent review), and in accordance are used for comparison in the experimental evaluation section: 1) softmax average, 2) simple voting, 3) weighted softmax average (where each softmax vector is multiplied by its related prediction confidence score), 4) confidence voting (where the most confident network gets $\frac{n}{2}$ votes), and 5) dictator voting (the decision of the most confident network prevails). We evaluate methods $3-5$ with weights defined by either the Entropy score or the Distance score defined in (\ref{eq:distscore}). 

\vspace{-.1cm}
\subsection{Novelty Detection}
\label{sec:new-novelty}

Novelty detection seeks to identify points in the test set which belong to classes not present in the train set. To evaluate performance in this task we train a network with a known benchmark dataset, while augmenting the test set with test points from another dataset that includes different classes. Each confidence score is used to differentiate between known and unknown samples. This is a binary classification task, and therefore we can evaluate performance using ROC curves. 

\section{Experimental Evaluation}
\label{sec:exp}

In this section we empirically evaluate the benefits of our proposed approach, comparing the performance of the new confidence score with alternative existing scores in the 3 different tasks described above.

\begin{table*}[t!]
\begin{center}
\caption{AUC Results of Correct Classification.}
~\\
\label{tbl:roc1}
\small{
\begin{tabular}{l|llll|llll|lll}
\hline
\multicolumn{1}{c|}{\begin{tabular}[c]{@{}c@{}}Conf.\\ Score\end{tabular}} & \multicolumn{4}{c|}{\begin{tabular}[c]{@{}c@{}}CIFAR-100 \\ (Classifier acccuracy: 60\%)\end{tabular}}               & \multicolumn{4}{c|}{\begin{tabular}[c]{@{}c@{}}STL-10 \\ (Classifier acccuracy: 70.5\%)\end{tabular}} & \multicolumn{3}{c}{\begin{tabular}[c]{@{}c@{}}SVHN\\ (Accuracy: 93.5\%)\end{tabular}} \\ \hline
                                                                               & \multicolumn{1}{c}{Reg.} & \multicolumn{1}{c}{Dist.} & \multicolumn{1}{c}{AT} & \multicolumn{1}{c|}{MCD} & Reg.                 & Dist.                & AT                   & MCD                 & Reg.                        & Dist.                      & AT                         \\ \hline
Margin                                                                     & 0.828                    & 0.834                     & 0.844                  & 0.836                   & 0.804                & 0.775                & 0.812                & 0.804               & 0.904                       & 0.896                      & 0.909                     \\ \hline
Entropy                                                                        & 0.833                    & 0.837                     & 0.851                  & 0.833                   & 0.806                & 0.786                & 0.816                & 0.810               & 0.916                       & 0.907                      & 0.918                      \\ \hline
Distance                                                                       & 0.789                    & \textbf{0.853}                     & 0.843                  & 0.726                   & 0.798                & 0.824                & \textbf{0.863}                & 0.671               & 0.916                       & \textbf{0.925}                      & \textbf{0.925}                      \\ \hline
\end{tabular}
}
\end{center}
~\\
\footnotesize{Table~\ref{tbl:roc1}, legend. Leftmost column: \textit{Margin} and \textit{Entropy} denote the commonly used confidence scores described in Section~\ref{sec:otherconf}. \textit{Distance} denotes our proposed method described in Section~\ref{sec:new-conf-score}. Second line: \textit{Reg.} denotes networks trained with the entropy loss, \textit{Dist.} denotes networks trained with the distance loss defined in (\ref{eq:dist-loss}), \textit{AT} denotes networks trained with adversarial training as defined in (\ref{eq:adverserial}), and \textit{MCD} denotes MC-Dropout when applied to networks normally trained with the entropy loss. Since the network trained for SVHN was trained without dropout,\textit{MCD} was not applicable.} 
\vspace{-0.1in}
\end{table*}

\begin{table*}[htb]
\begin{center}
\caption{AUC Results of Correct classification - Ensemble of 2 Networks.}
~\\
\label{tbl:roc2}
\begin{tabular}{l|lll|lll|lll}
\hline
\multicolumn{1}{c|}{\begin{tabular}[c]{@{}c@{}}Confidence\\ Score\end{tabular}} & \multicolumn{3}{c|}{CIFAR-100}                                                 & \multicolumn{3}{c|}{STL-10} & \multicolumn{3}{c}{SVHN} \\ \hline
                                                                               & \multicolumn{1}{c}{Reg.} & \multicolumn{1}{c}{Dist.} & \multicolumn{1}{c|}{AT} & Reg.    & Dist.   & AT     & Reg.   & Dist.  & AT     \\ \hline
Max margin                                                                     & 0.840                    & 0.846                     & 0.856                  & 0.802   & 0.792   & 0.814  & 0.909  & 0.901  & 0.911  \\ \hline
Entropy                                                                        & 0.844                    & 0.839                     & 0.857                  & 0.807   & 0.798   & 0.816  & 0.918  & 0.912  & 0.920  \\ \hline
Distance (1)                                                                   & 0.775                    & 0.863                     & 0.862                  & 0.815   & 0.834   & \textbf{0.866}  & 0.916  & 0.924  & 0.926  \\ \hline
Distance (2)                                                                   & 0.876                    & 0.872                     & \textbf{0.879 }                 & 0.833   & 0.832   & 0.859  & 0.918  & \textbf{0.929}  & 0.927 \\ \hline
\end{tabular}
\end{center}
~\\
\footnotesize{Table~\ref{tbl:roc2}, legend. Notations are similar to those described in the legend of Table~\ref{tbl:roc1}, with one distinction: \textit{Distance (1)} now denotes the regular architecture where the distance score is computed independently for each network in the pair using its own embedding, while \textit{Distance (2)} denotes the hybrid architecture where one network in the pair is fixed to be a \textit{Distance} network, and its embedding is used to compute the distance score for the prediction of the second network in the pair. }
\vspace{-0.1in}
\end{table*}

\subsection{Experimental Settings}

For evaluation we used 3 data sets: CIFAR-100 \citep{krizhevsky2009learning}, STL-10 \citep{coates2010analysis} ($32\times 32$ version) and SVHN \citep{netzer2011reading}. In all cases, as is commonly done, the data was pre-processed using global contrast normalization and ZCA whitening. No other method of data augmentation was used for CIFAR-100 and SVHN, while for SVHN we also did not use the additional \~500K labeled images\footnote{Note that reported results denoted as "state-of-the-art" for these datasets often involve heavy augmentation. In our study, in order to be able to do the exhaustive comparisons described below, we opted for the un-augmented scenario as more flexible and yet informative enough for the purpose of comparison between different methods. Therefore our numerical results should be compared to empirical studies which used similar \emph{un-augmented settings}. We specifically selected commonly used architectures that achieve good performance, close to the results of modern ResNets, and yet flexible enough for extensive evaluations.}. For STL-10, on the other hand, cropping and flipping were used for STL-10 to check the robustness of our method to heavy data augmentation.

\comment{We first describe the neural networks used in the following experiments for the respective datasets, including the hyper parameters used during training. All networks were trained using the TensorFlow environment with Exponential-Linear-Unit activations (ELU) \citep{clevert2015fast}, augmented with an $L_2$ regularizer with weight decay factor of $10^{-4}$. In all experiments we used batch size of 100.} 

In our experiments, all networks used ELU \citep{clevert2015fast} for non-linear activation.  For CIFAR-100 and STL-10 we used the network suggested in \cite{clevert2015fast} with the following architecture:

$C(192,5)\Rightarrow P(2)\Rightarrow C(192,1)\Rightarrow C(240,3) \Rightarrow P(2) \Rightarrow C(240,1)\Rightarrow C(260,3) \Rightarrow P(2) \Rightarrow C(260,1)\Rightarrow C(280,2) \Rightarrow P(2) \Rightarrow C(280,1)\Rightarrow C(300,2) \Rightarrow P(2) \Rightarrow C(300,1) \Rightarrow FC(100)$ 

$C(n,k)$ denotes a convolution layer with $n$ kernels of size $k\times k$ and stride 1. $P(k)$ denotes a max-pooling layer with window size $k\times k$ and stride 2, and $FC(n)$ denotes a fully
connected layer with $n$ output units. For STL-10 the last layer was replaced by FC(10). During training (only) we applied dropout \citep{srivastava2014dropout} before each max pooling layer (excluding the first) and after the last convolution, with the corresponding drop probabilities of $[0.1, 0.2, 0.3, 0.4, 0.5]$.  \comment{Note that the last convolution layer output feature map size is $ 1 \times 1$, and thus, after reshaping, it can be treated as an embedding vector of size 300.}

\comment{We trained the first network using momentum optimizer \cite{qian1999momentum} for 80,000 steps and with the following learning rates: [steps 0-30K: $10^{-2}$, steps 30K-50K: $5*10^{-3}$, steps 50K-65K: $5*10^{-4}$, steps 65K-80K: $5*10^{-5}$].} 

\comment{\paragraph{STL-10} 
The network and all parameters are the same as in CIFAR-100. The only differences are the last FC layer, which is now of size 10 (since we have 10 classes), and the fact that we train for 8K steps only using the following learning rates:  [steps 0-3K: $10^{-2}$, steps 3K-5K: $5*10^{-3}$, steps 5K-6.5K: $5*10^{-4}$, steps 6.5K-8K: $5*10^{-5}$]. }

With the SVHN dataset we used the following architecture: 

$C(32,3)\Rightarrow C(32,3) \Rightarrow C(32,3) \Rightarrow P(2) \Rightarrow C(64,3)\Rightarrow C(64,3)\Rightarrow C(64,3)\Rightarrow P(2) \Rightarrow C(128,3)\Rightarrow C(128,3)\Rightarrow C(128,3)\Rightarrow P(2) \Rightarrow FC(128) \Rightarrow FC(10)$ 

\comment{For both datasets we trained the first network using the ADAM \cite{kingma2014adam} optimizer. For SVHN we trained for 16K steps with the following learning rates: [steps 0-6K: $10^{-3}$, steps 6K-10K: $5*10^{-4}$, steps 10K-16K: $5*10^{-5}$]. For MNIST we trained for 6K steps with following learning rates: [steps 0-2.5K: $10^{-3}$, steps 2.5K-4K: $5*10^{-4}$, steps 4K-6K: $5*10^{-5}$].}

For the networks trained with distance loss, for each batch, we randomly picked pairs of points so that at least 20\% of the batch included pairs of points from the same class. The margin $m$ in ~(\ref{eq:dist-loss}) was set to 25 in all cases, and the parameter $\alpha$ in ~(\ref{eq:dist-loss}) was set to 0.2. The rest of the training parameters can be found in the supplementary material. For the distance score we observed that the number of $k$ nearest neighbors could be set to the maximum value, which is the number of samples in each class in the train data. We also observed that smaller numbers (even $k=50$) often worked as well. In general, the results reported below are not sensitive to the specific values of the hyper-parameters as listed above; we observed only minor changes when changing the values of $k, \alpha$ and the margin $m$.

~\\
\emph{MC-Dropout:} 
as proposed in \cite{gal2015dropout}, we used MC dropout in the following manner. We trained each network as usual, but computed the predictions while using dropout during test. This was repeated 100 times for each test example, and the average activation was delivered as output.

\emph{Adversarial Training:} we used (\ref{eq:adverserial}) following \cite{goodfellow2014explaining}, fixing $\epsilon=0.1$ in all the experiments.

\begin{figure*}[!hbt] 
	\centering
	\includegraphics[width=0.9\textwidth, height=0.2\textheight]{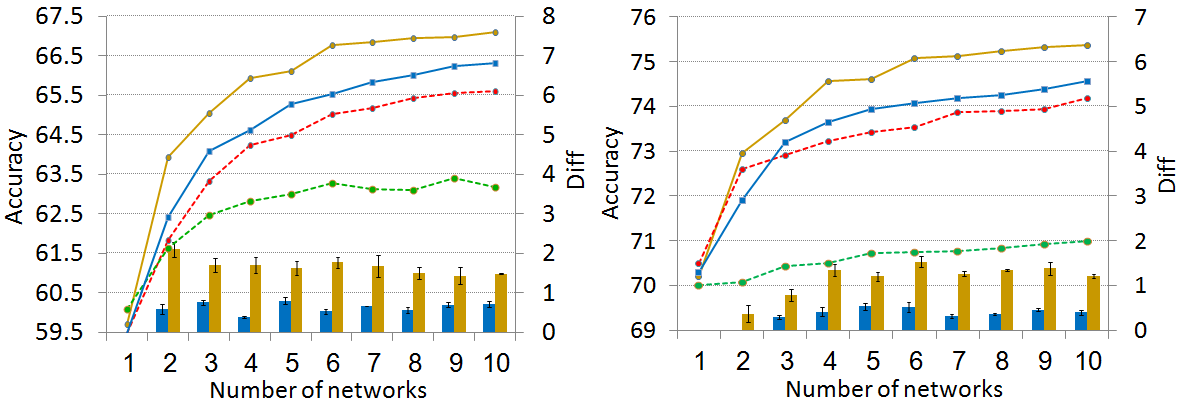}
	~\\~\\
            	\includegraphics[width=0.45\textwidth, height=0.2\textheight]{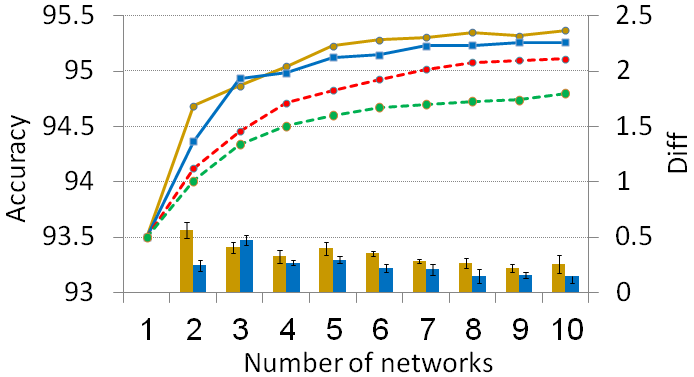}
                    	\raisebox{0.23\height}{\includegraphics[width=0.45\textwidth, height=0.15\textheight]{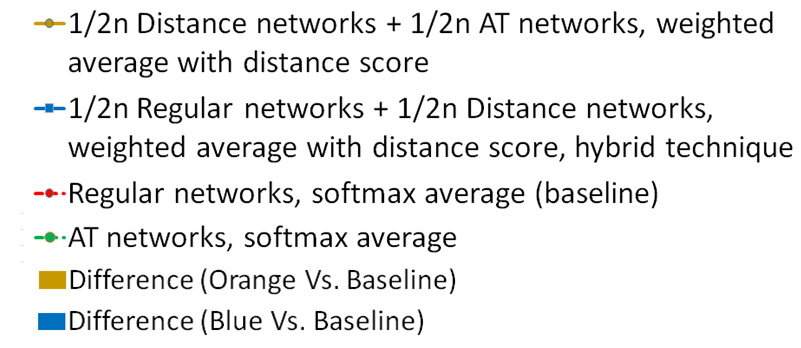}}
	\caption{Accuracy when using an ensemble of networks with CIFAR-100 (top left), STL-10 (top right) and SVNH (bottom). The $X$-axis denotes the number of networks in the ensemble. Absolute accuracy (marked on the left $Y$-axis) is shown for the 2 most successful ensemble methods among all the methods we had evaluated (blue and yellow solid lines, see text), and 2 methods which did not use our distance score including the best performing method in this set (red dotted line, denoted baseline). Differences in accuracy between the two top performers and the top baseline method are shown using a bar plot (marked on the right $Y$-axis), with standard deviation of the difference over (at least) 5 repetitions. 
    } 
    \label{fig:cifar100en}
\end{figure*}

\subsection{Error Prediction of Multi-class Labels} 
\label{sec:exproc}
We first compare the performance of our confidence score in the binary task of evaluating whether the network's predicted classification label is correct or not. While our results are independent of the actual accuracy, we note that the accuracy is comparable to those achieved with ResNets when not using augmentation for CIFAR-100 or when using only the regular training data for SVHN (see \cite{huang2016deep} for example). 

Performance in this binary task is evaluated using ROC curves computed separately for each confidence score. Results on all three datasets can be seen in Table~\ref{tbl:roc1}. In all cases our proposed distance score, when computed on a suitably trained network, achieves significant improvement over the alternative scores, even when those are enhanced by using either Adversarial Training or MC-Dropout. 

To further test our \emph{distance score} we evaluate performance over an ensemble of two networks. Results are shown in Table~\ref{tbl:roc2}. Here too, the distance score achieves significant improvement over all other methods. We also note that the difference between the distance score computed over \emph{Distance networks} and the entropy score computed over adversarially trained networks, is now much higher as compared to this difference when using only one network. As we show in Section~\ref{sec:expensemble}, adversarial training typically leads to a decreased performance when using an ensemble of networks and relying only on the entropy score (probably due to a decrease in variance among the classifiers). This observation further supports the added value of our proposed confidence score.

As a final note, we also used a hybrid architecture using a matched pair of one classification network (of any kind) and a second \emph{Distance network}. The embedding defined by the \emph{Distance network} is used to compute the distance score for the predictions of the first classification network. Surprisingly, this method achieves the best results in both CIFAR-100 and SVHN while being comparable to the best result in STL-10. This method is used later in Section~\ref{sec:expensemble} to improve accuracy when running an ensemble of networks. Further investigation of this phenomenon lies beyond of the scope of the current study.

\subsection{Ensemble Methods}
\label{sec:expensemble}
\vspace{-0.05in}
In order to evaluate the improvement in performance when using our confidence score to direct the integration of classifiers in an ensemble, we used a few common ways to define the integration procedure, and a few ways to construct the ensemble itself. In all comparisons the number of networks in the ensemble remained fixed at $n$. Our experiments included the following ensemble compositions: (a) $n$ \emph{regular networks}, (b) $n$ \emph{distance networks}, (c) $n$ \emph{AT (Adversarially Trained) networks}, and (d-f) $n$ networks such that $\frac{n}{2}$ networks belong to one kind of networks (\emph{regular}, \emph{distance} or \emph{AT}) and the remaining $\frac{n}{2}$ networks belong to another kind, spanning all 3 combinations.

As described in Section~\ref{sec:new-ensemble}, the predictions of classifiers in an ensemble can be integrated using different criteria. In general we found that all the methods which did not use our distance score (\ref{eq:distscore}), including methods which used any of the other confidence score for prediction weighting, performed less well than a simple average of the softmax activation (method 1 in Section~\ref{sec:new-ensemble}). Otherwise the best performance was obtained when using a weighted average (method 3 in Section~\ref{sec:new-ensemble}) with weights defined by our distance score (\ref{eq:distscore}). With variants (d-f) we also checked two options of obtaining the distance score: (i) Each network defined its own confidence score; (ii) in light of the advantage demonstrated by hybrid networks as shown in Section~\ref{sec:exproc} and for each pair of networks from different kinds, the distance score for both was computed while using the embedding of only one of networks in the pair. MC-Dropout was not used in this section due to its high computational cost.

While our experiments included all variants and all weighting options, only 4 cases are shown in the following description of the results in order to improve readability: 1) the combination achieving best performance; 2) the combination achieving best performance when not using adversarial training (as \emph{AT} entails additional computational load at train time); 3) the ensemble variant achieving best performance without using the distance score (baseline), 4) ensemble average when using adversarial training without distance score. Additional results for most of the other conditions we tested can be found in the supplementary material. To gain a better statistical significance, each experiment was repeated at least 5 times, with no overlap between the networks.  

\textbf{CIFAR-100 and STL-10:} Fig.~\ref{fig:cifar100en} shows the  ensemble accuracy for the methods mentioned above when using these datasets. It can be clearly seen that weighting predictions based on the distance score from (\ref{eq:distscore}) improves results significantly. The best results are achieved when combining \emph{Distance Networks} and \emph{Adversarial Networks}, with significant improvement over an ensemble of only one kind of networks (not shown in the graph). Still, we note importantly that the distance score is used to weight \textbf{both} kind of networks. Since Adversarial Training is not always applicable due to its computational cost at train time, we show that the combination of \emph{Distance networks} and \emph{Regular networks} can also lead to significant improvement in performance when using the distance score and the hybrid architecture described in Section~\ref{sec:exproc}. Finally we note that \emph{Adversarial networks} alone achieve very poor results when using the original ensemble average, further demonstrating the value of the distance score in improving the performance of an ensemble of Adversarial networks alone.  

\textbf{SVHN:} Results for this dataset are also shown in Fig.\ref{fig:cifar100en}. While not as significant as those in the other datasets (partly due to the high initial accuracy), they are still consistent with them, demonstrating again the power and robustness of the distance score. 

\vspace{-0.12in}
\subsection{Novelty Detection}
\vspace{-0.02in}
Finally, we compare the performance of the different confidence scores in the task of novelty detection. In this task the confidence score is used to decide another binary classification problem: does the test example belong to the set of classes the networks had been trained on, or rather to some unknown class? Performance in this binary classification task is evaluated using the corresponding ROC curve of each confidence score. 

We used two contrived datasets to evaluate performance in this task, following the experimental construction suggested in \cite{lakshminarayanan2016simple}. In the first experiment, we trained the network on the STL-10 dataset, and then tested it on both STL-10 and SVHN test sets. In the second experiment we switched between the datasets (and changed the trained network) making SVHN the known dataset and STL-10 the novel one. The task requires to discriminate between the known and the novel datasets. For comparison we computed novelty, as one often does, with a one-class SVM classifier while using the same embeddings. Novelty thus computed showed much poorer performance, possibly because this dataset involves many classes (one class SVM is typically used with a single class), and therefore these results are not included here. 

\begin{table}[th!]
\begin{center}
\caption{AUC Results for Novelty Detection. }
\label{tbl:novel}
~\\
{\small
\begin{tabular}{l|lll|lll}
\hline
\multicolumn{1}{c|}{\begin{tabular}[c]{@{}c@{}}Confide.\\ Score \end{tabular}} & \multicolumn{3}{c|}{STL-10/SVHN}                                               & \multicolumn{3}{c}{SVHN/STL-10} \\ \hline
                                                                               & \multicolumn{1}{c}{Reg.} & \multicolumn{1}{c}{Dist.} & \multicolumn{1}{c|}{AT} & Reg.      & Dist.    & AT       \\ \hline
\begin{tabular}[c]{@{}c@{}}Max \\ Margin\end{tabular}  & .808                   & .849                     & .860                   & .912    & .922   & .985   \\ \hline
Entropy                                                                        & .810                   & .857                    & .870                 & .917    & .933   & .992   \\ \hline
Distance                                                                       & .798                   & .870                    & \textbf{.901}                  & .904    & .934   & \textbf{.996}   \\ \hline
\end{tabular}
}
\end{center}
~\\
\footnotesize{Table~\ref{tbl:novel}, legend. Left: STL-10 (known) and SVHN (novel). Right: SVHN (known) and STL-10 (novel).}
\end{table}

Results are shown in Table~\ref{tbl:novel}. Adversarial training, which was designed to handle this sort of challenge, is not surprisingly the best performer. Nevertheless, we see that our proposed confidence score improves the results even further, again demonstrating its added value. 

\section{Conclusions}

We proposed a new confidence score for multi-class neural network classifiers. The method we proposed to compute this score is scalable, simple to implement, and can fit any kind of neural network. This method is different from other commonly used methods as it is based on measuring the point density in the effective embedding space of the network, thus providing a more coherent statistical measure for the distribution of the network's predictions.  
 
We also showed that suitable embeddings can be achieved by using either a distance-based loss or, somewhat unexpectedly, Adversarial Training.  We demonstrated the superiority of the new score in a number of tasks. Those tasks were evaluated using a number of different datasets and with task-appropriate network architectures. In all tasks our proposed method achieved the best results when compared to traditional confidence scores.


\bibliography{paperbib}
\bibliographystyle{icml2017}

\end{document}